\pdfoutput=1

\documentclass[11pt]{article}

\usepackage[]{EACL2023}

\usepackage{times}
\usepackage{latexsym}

\usepackage[T1]{fontenc}

\usepackage[utf8]{inputenc}

\usepackage{microtype}

\usepackage{inconsolata}
\usepackage{multirow}
\usepackage{makecell}
\usepackage{algorithm}
\usepackage{algorithmicx}
\usepackage{booktabs}
\usepackage{algpseudocode}
\usepackage{algcompatible}
\usepackage{amsfonts}
\usepackage{amsmath}
\usepackage{caption}
\usepackage{subcaption}
\usepackage{hyperref} 
\usepackage{xspace}
\usepackage{enumitem}
\usepackage{tikz}
\usepackage{pgfplots}
\usepackage{arydshln}
\usepackage{soul}
\usepackage{xcolor}
\usepackage{mathabx}
\usepackage{balance}
\usetikzlibrary{positioning}
\pgfplotsset{compat=1.15}

\usepackage[textsize=scriptsize]{todonotes}

\newcommand{\gd}[1]{\textcolor{red}{Greg:[#1]}}

\newcommand{\eat}[1]{}

\newcommand{\qgelm}{question-guided event language model\xspace}
\newcommand{\e}{EGELM\xspace}
\newcommand{\q}{QGELM\xspace}
\newcommand{\base}{ELM\xspace}

\title{{M}odeling {C}omplex {E}vent {S}cenarios via {S}imple {E}ntity-focused {Q}uestions}

\author{Mahnaz Koupaee$^1$, Greg Durrett$^2$, Nathanael Chambers$^3$, Niranjan Balasubramanian$^1$ \\
\\
  $^1$ Stony Brook University,
  $^2$ The University of Texas at Austin,
  $^3$ United States Naval Academy\\
  $^1$\texttt{\{mkoupaee,niranjan\}@cs.stonybrook.edu}\\
  $^2$\texttt{gdurrett@cs.utexas.edu},
  $^3$\texttt{nchamber@usna.edu}\\}

\begin{document}
\maketitle
\begin{abstract}\label{seq:abs}

Event scenarios are often complex and involve multiple event sequences connected through different entity participants. Exploring such complex scenarios requires an ability to branch through different sequences, something that is difficult to achieve with standard event language modeling. To address this, we propose a question-guided generation framework that models events in complex scenarios as answers to questions about participants.
At any step in the generation process, the framework uses the previously generated events as context, but generates the next event as an answer to one of three questions: \textit{what else a participant did, what else happened to a participant, or what else happened}. The participants and the questions themselves can be sampled or be provided as input from a user, allowing for controllable exploration. Our empirical evaluation shows that this question-guided generation provides better coverage of participants, diverse events within a domain, comparable perplexities for modeling event sequences, and more effective control for interactive schema generation\footnote{The code is available at \url{https://github.com/StonyBrookNLP/qa-event-lms}}.

\end{abstract}
\section{Introduction}\label{sec:intro}

Event scripts \cite{schank1977scripts}, also known as event schemas, describe a sequence of events in a particular context. Representing and modeling such schemas is central to applications in AI such as question answering, discourse understanding, and information extraction \cite{balasubramanian2013generating}. Early work used hand-crafted event schemas as a starting point \cite{schank1977scripts,mooney1985learning}, but modern techniques attempt to extract these at a large scale from unlabeled data \cite{chambers2008unsupervised,chambers2013event,pichotta2016learning, weber2018hierarchical}. 

%
Event language models can also be used to approximate schematic knowledge via event sequences. They can be trained to generate a sequence of events with their participating roles describing a real-life scenario.
However, these scenarios can often be complex and don't always fit as simple sequences \cite{weber2018hierarchical}. For example, suppose we have the following event: \texttt{police arrested suspect on several charges}, with \textbf{\texttt{police}}, \textbf{\texttt{suspect}} and \textbf{\texttt{charges}} as entities.
\begin{figure}
\centering
\includegraphics[width=0.4\textwidth]{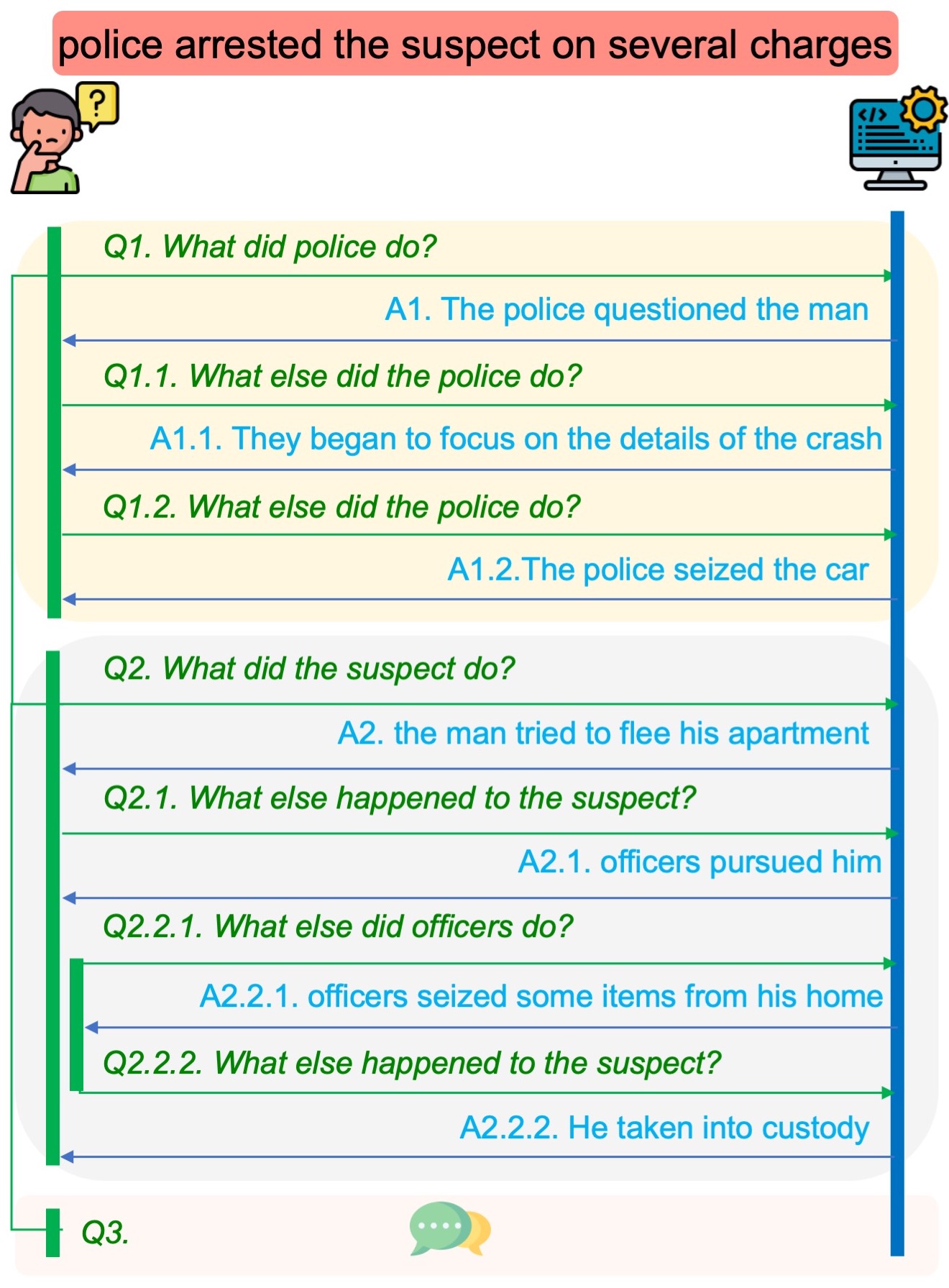}
\caption{Question-guided event sequence generation. The user can interact with the system by asking questions regarding the entities and the system will generate corresponding events. Different questions can lead to different paths as shown with different colored boxes.}
\label{fig:overview}
\end{figure}
The scenario can be described in multiple ways depending on which entities we want to focus on and what roles they play in the subsequent events. We may be interested in knowing what the \textbf{\texttt{police}} did, what the \textbf{\texttt{suspect}} did or what happened to the \textbf{\texttt{charges}}, each of which can be explored as its own sequence, yielding many interconnected sequences of events in the scenario. An event language model, however, will simply generate events conditioned on the previous events in the discourse, with no direct mechanism to guide generation towards areas of interest within the scenario. As a result, with standard decoding strategies we often end up with a sequence of events that might be relevant to the scenario, but not necessarily cover the broad set of diverse paths in the scenario; we would only know the fate of certain entities if the system samples events that included them.
This lack of control will make it difficult to use the system, as one must keep sampling events until it just happens to produce events with roles one is interested in.

We propose a simple modification in which event language models are trained to also condition on the entities and the roles they play in a scenario through a set of simple questions, as illustrated in \hyperref[fig:overview]{Figure 1}. Given the same example event and its entities, one can then explore the scenario through various paths by asking the system to generate the events with their desired entities and roles. Such a question-guided model can be used in interactive settings to model and construct diverse paths that cover various aspects of complex scenarios. 

A key challenge, however, is in creating the necessary training data at scale. We show that we can repurpose standard event sequences to create training data for question-guided models: we take a partial sequence of events as context and derive a role-based question involving an entity for which a future event in the sequence can be an answer. This allows for creating large scale training instances that are (Context, Question, Answer) triples, which then allows us to train models that can better respond to user control in the form of questions.

Our analysis shows that question-guided event language models can generate sequences with more diversity and comparable quality as an event language model. Our human evaluation of the model in an interactive setting, shows that the controllability of the question-guided model allows for generation of sequences that lead to better quality, broader-coverage schemas with fewer interactions. 
This interactive evaluation is a step toward constructing schematic/common-sense knowledge for analyzing events, an application in the intelligence analysis community.

In summary, this paper makes the following contributions; (i) It argues the need for control in event language models to explore complex scenarios; (ii) It provides a simple yet effective way for training event language models that can be guided to explore different aspects of complex scenarios; (iii) It provides empirical evidence showing improved control and utility via automatic and manual evaluations.



\section{Related Work}\label{sec:rel}
\textbf{Event Schema Induction} Event scripts (or schemas) originally proposed by \citep{schank1977scripts}, consist of a set of events and actors (also known as slots) playing different roles. The event schemas are capable of analyzing complex situations by encoding information from prototypical events and their participants. 

Early works on scripts considered them as structured representations of events and their participants with the causal relationships between them \citep{schank1977scripts,mooney1985learning}.
However, the manual construction of scripts is too time-consuming and does not scale, so the scripts could only focus on specific domains of interest and have not been used more broadly.

Event schemas can also be induced automatically from text using statistical techniques in an unsupervised fashion \cite{chambers2008unsupervised,chambers2009unsupervised, balasubramanian2013generating}. These models are easily interpretable but fail to capture long-distance complex relationships between events. 
Event language modeling is a type of schema induction via language modeling techniques \citep{rezaee2021event}. Given a sequence of events, the event language model predicts the probability of the next event \cite{manshadi2008learning}.
Framing schema learning as a language modeling problem with various ways to represent events, including word sequences annotated with predicate-argument structure \cite{pichotta2016learning}, OpenIE tuples \cite{rudinger2015script,weber2018event,weber2018hierarchical} or compositional embeddings \cite{modi2016event}, is another direction towards realizing large-scale schema libraries. 

Graph schema induction methods \cite{li-etal-2020-connecting,li2021future} model different relations between entities and their arguments to capture the multi-dimensionality of scenarios.
Reasoning about complex relations between events requires going beyond the single dimension of event cooccurrence and capturing different types of semantic relations between events such as causal, counterfactual, etc. \cite{han-etal-2021-ester}.  
Our approach uses the standard language models and provides them with the guidance to also produce different aspects of a scenario via simple control codes.


\noindent \textbf{Controlled text generation} Language models have shown promising results in text generation, however, it is not easy to have control over different aspects of generation \cite{keskar2019ctrl}, an issue that has been studied in previous works \cite{dathathri2019plug,he2020ctrlsum,lu-etal-2021-neurologic,mireshghallah-etal-2022-mix}.
The key components of these methods differ in the types of controls provided and how they are provided and their applications.
CTRL \cite{keskar2019ctrl} trains a very large language model by conditioning on texts with appended control codes that are used to guide the generation towards specific styles, contents, and task-specific behaviors. 
CTRLSum \cite{he2020ctrlsum} uses the entity/length controls which are in forms of keywords that are automatically extracted from the text and trains a summarization system which is capable of generating summaries in an interactive manner.
These approaches, however, require finetuning. \citet{dathathri2019plug} and \citet{mireshghallah-etal-2022-mix} propose variants of controllable generation with no need to finetune or retrain the whole system.

Learning latent representations or codes from the input towards diverse generation is another direction that has been explored for machine translation \cite{shu-etal-2019-generating}, dialogue generation \cite{huang-etal-2018-automatic} and causal relations generation \cite{weir-etal-2020-cod3s}.
Controllable generation to generate diverse events has been previously studied in \citet{kwon2021toward}, where the system uses automatically generated control codes to generate diverse preconditions of events.

In this work, we use controllable generation to model complex event scenarios. We introduce simple role-based questions about participants (agentive or non-agentive) as an effective means for control. 
Asking questions to get specific information about events is the focus of many existing approaches. While there has also been a body of work on semantic role labeling using QA pairs \cite{roit-etal-2020-controlled,klein-etal-2020-qanom,michael-zettlemoyer-2021-inducing,pyatkin-etal-2021-asking}, the main distinction here lies in the fact that these approaches use QA pairs to identify the semantic roles, whereas our approach makes use of role-based questions to generate the next event with a specific entity playing a specific role.
We show how to train for these control codes using automatically derived training sequences and demonstrate its utility for describing complex scenarios in an interactive setting.

\begin{figure}
\centering
\includegraphics[width=0.44\textwidth]{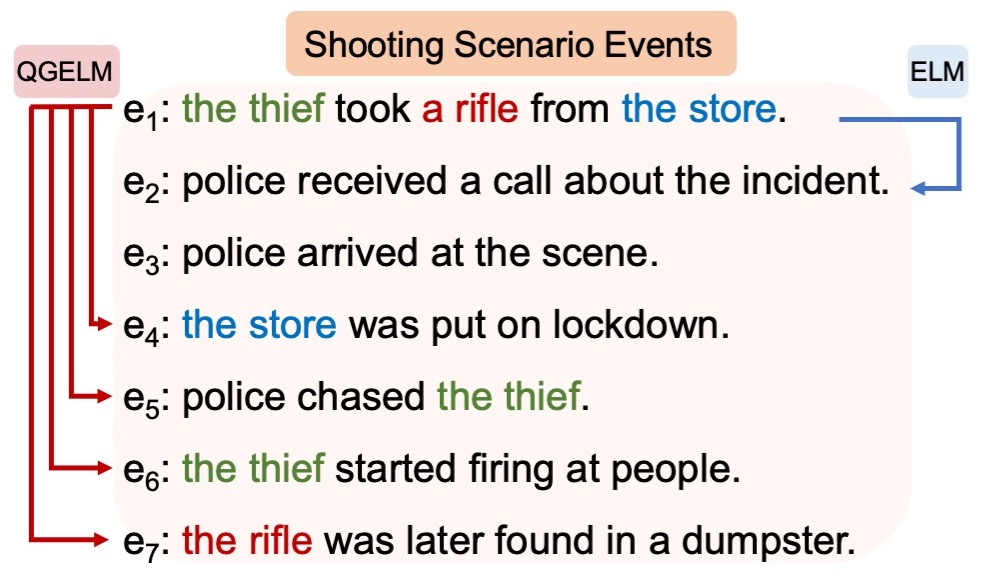}
\caption{The shooting scenario events extracted from a news article. Typical LMs only see $e_2$ as the immediate next event, whereas for the question-guided LMs, any of the $e_4$, $e_5$, $e_6$ and $e_7$ involving \textit{the store}, \textit{the thief}, or \textit{the rifle} can be considered a next event.}
\label{fig:motiv}
\end{figure}


\section{Question-guided Event Language Modeling}\label{sec:method}

\begin{figure*}
\centering
\includegraphics[width=\textwidth]{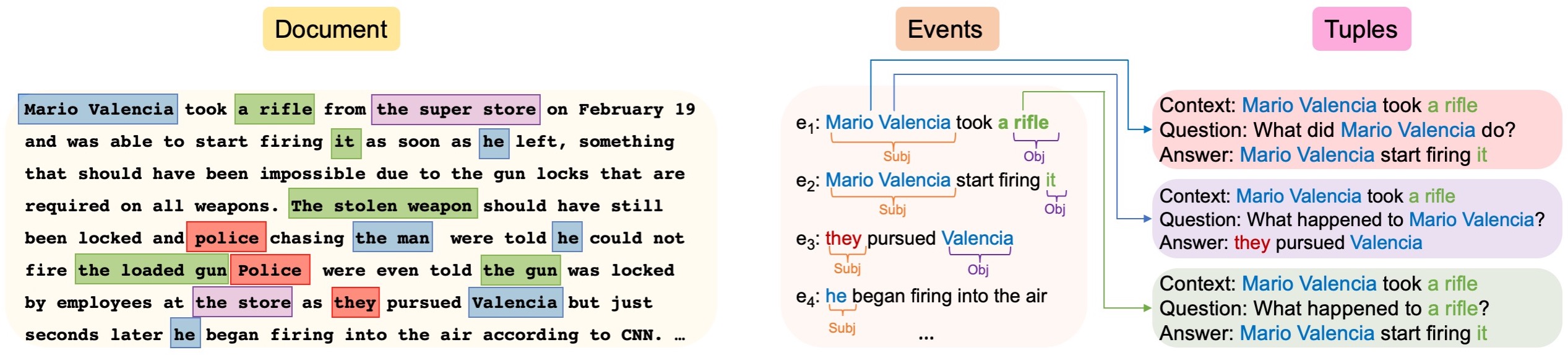}
\caption{Data processing to create instances for a \qgelm. Given a document, we extract all the coreferring clusters (color-coded in the document). Next, we extract OpenIE tuples as events representations. By identifying the roles of noun phrases in the events as well as knowing which cluster they belong to, we create questions for each one of them and similarly find the events that can serve as the answers to those questions. The context can be of any length whereas the answer event is of length $1$.}
\label{fig:data}
\end{figure*}

Event language models aim at predicting the probability of an event given a set of events via a conditional probability distribution. 
Formally, these models try to find an event $\hat{e}$ by maximizing the probability of a function parameterized by a model over a given context including a set of events:

\[ \mathrm{\hat{e} = \mathrm{argmax}_{e \in E} P_\theta(e \mid \mathrm{context})}\]

\noindent where $E$ is the set of all events. A model trained with this objective learns to generate the most probable event based on a set of cooccurring events from discourse. 
Suppose that shooting scenario events are extracted as shown in \hyperref[fig:motiv]{Figure 2}.
Given the first event in the sequence, a typical language model is trained to generate the next event, $e_2$, which does not include any entities from the prior context. 
What if we are interested in looking for events regarding the initial participants (\textit{the thief}, \textit{a rifle}, or \textit{the store}) in this example? Given the same context, we can have multiple next events depending on the participant and the role they can have (agent or theme). 
If we are interested in knowing what \textit{the thief} did, $e_6$ should be the next event, what happened to \textit{the thief} is described in $e_5$. We can have some information about \textit{the store} with $e_4$ as the next event and finally, the fate of \textit{the rifle} is described in $e_7$.
Controlling the model to generate the outputs based on the participants involved cannot be easily achieved with unguided (text ordered) event language models. However, we will show how the language models can be guided such that given the same context, they can directly generate events for its participants.



\subsection{Problem Definition}
We propose a new framing in which an event language model can be guided to generate events by not only conditioning on the events but also on specific entity-based questions of interest. 

These entity-aware event language models also look to find $\hat{e}$, but by maximizing the probability of a function parameterized by a model over a given context \emph{and} a question regarding an entity:
\[ \mathrm{\hat{e} = \mathrm{argmax}_{e \in E} P_\theta(e \mid \mathrm{context}, \mathrm{question})}\]
The objective now conditions on a question as well, and the goal is to use the question to train a system to generate the most probable event for a specific entity (or a noun phrase referring to that entity) in a specific role (agent or theme). 
As shown in \hyperref[fig:overview]{Figure 1}, questions can either ask about what an entity did as an agent of an action (\textit{what did X do?}) or what happened to an entity as a theme of an action (\textit{what happened to X?}). 
By conditioning on the question as well, the system will learn to generate an event with the entity in question as well as the role specified by that question.

\subsection{Question-guided Training}\label{sec:training}

Event language models are typically trained using event sequences extracted from documents. Our goal, however, is to train event language models to generate events as \textit{answers} to \textit{questions} about entities from a given \textit{context}. To this end, we convert event sequences in text to \textit{(Context, Question, Answer)} tuples (CQA instances) that can be used as training data for question-guided generation. 

Consider a sequence $e_1,e_2,...e_n$ of OpenIE event tuples extracted from a document $D$. We create \textit{(Context, Question, Answer)} tuples for each event $e_t$ in the sequence as outlined in Algorithm~\ref{alg:migrate} in Appendix~\ref{append:data} and \hyperref[fig:data]{Figure 3}. The key idea behind this process is as follows: Suppose we observe an entity in an event $e_i$. If this entity also appears in a subsequent event $e_k$, then we can see this new event as an answer to a role-based question about the entity (what did the entity do or what happened to the entity), given what we know about all the events that have been observed thus far in the sequence as context.
For example, as shown in \hyperref[fig:data]{Figure 3}, the entity  \texttt{Mario Valencia} appears in two events $e_1$ and $e_2$. Given the context $e_1$, we can create the question \textit{What did Mario Valencia do?} for which the answer is $e_2$ i.e., \textit{Mario Valencia started firing it}. 
Formally, for each entity (any noun phrase) $np$ that appears as an argument in the event $e_i$, we do the following. If $np$ appears in an agentive role in some subsequent event $e_{k}$ ($k>i$), then we associate the question \textit{What else did $np$ do?} with the context and use event $e_{k}$ as the answer to the question. If $np$ appears in a non-agentive role, we associate \textit{What else happened to $np$?} as the question
and $e_{k}$ as the answer. In either case, for the next step the context is  extended to include $e_{k}$ and the process is repeated for all arguments in $e_{k}$. To handle events $e_{k}$ that introduce new entities as arguments, we use \textit{What else happened?} as the question.  

We use automatically identified coreference clusters to locate event mentions involving a specific entity. We use simple dependency-based heuristics to determine the role of an entity in an event. Given the noun phrases of an event, we use a dependency parser to identify their roles. 
An entity is deemed to appear in an agentive role if it appears as a subject and in a non-agentive role if it appears as an object. We only use these two broad categories (subject and object) to identify events as responses to specific questions.
This process will automatically filter out the nonsensical questions as for such questions, no event can be found within the given sequence of events.

Note this training data has two properties that are different from standard auto-regressive training over event sequences. First, for the same conditioning context of event tuples, the model learns to generate multiple subsequent events depending on the question being asked, thus ensuring better control and diversity. Second, the model also learns to generate events that are not always adjacent to the end of the current sequence, which can be seen as a form of data augmentation shown to be effective \cite{koupaee-etal-2021-dont}.
\section{Experimental Setup}

\begin{figure}[t]
\centering
\large
\begin{tikzpicture}[scale=0.8]
\begin{axis}[
    xlabel={Sequence length},
    ylabel={Self-BLEU3},
    xmin=0, xmax=11,
    ymin=6, ymax=19,
    xtick={1,2,3,4,5,6,7,8,9,10},
    ytick={6,8,10,12,14,16,18},
    legend pos=north west,
]

\addplot[
    color=teal,
    mark=square,
    mark size=3pt,
    line width=0.3mm,
    ]
    coordinates {
  (1,7.22)(2,8.55)(3,10.08)(4,11.5)(5,12.61)(6,13.57)(7,14.9)(8,15.7)(9,16.63)(10,17.14)};
 
\addplot[
    color=blue,
    mark=pentagon,
    mark size=3pt,
    line width=0.3mm,
    ]
    coordinates {
  (1,7.35)(2,8.64)(3,9.84)(4,10.71)(5,12.03)(6,13.28)(7,13.86)(8,14.67)(9,15.53)(10,16.26)};

\addplot[
    color=magenta,
    mark=otimes,
    mark size=3pt,
    line width=0.3mm,
    ]
    coordinates {
  (1,6.83)(2,7.51)(3,8.44)(4,9.6)(5,10.18)(6,11.08)(7,11.9)(8,12.81)(9,13.54)(10,14.1)};

\addlegendentry{\base}
\addlegendentry{\e}
\addlegendentry{\q}
 
\end{axis}
\end{tikzpicture}
\caption{Diversity of generated sequences of events with varying lengths. Lower Self-BLEU scores are better as they represent more diverse sequences. \q is the most diverse across all sequence lengths.}
  \label{fig:diversity}
\end{figure}
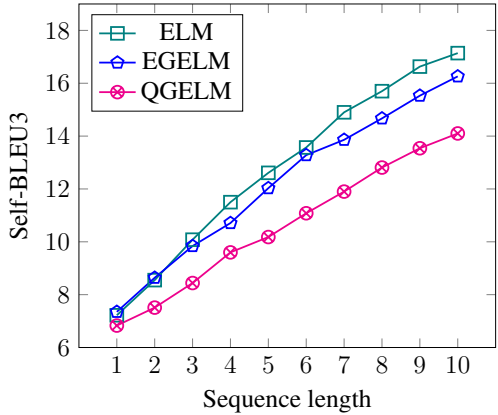

\subsection{Models}
All models used in our experiments are trained using event sequences, which are OpenIE~\cite{ollie-emnlp12} tuples extracted from articles in their discourse order. 

\paragraph{\base}
Our baseline is an \textbf{E}vent \textbf{L}anguage \textbf{M}odel trained such that given a 
context (a set of events), it will generate the next event. 
The baseline model follows the existing mechanisms used in recent prior ELM work \cite{manshadi2008learning, rudinger2015script, pichotta2016learning, weber2018event, weber2018hierarchical}.
We train a T5 base model to learn \( P(e_n \mid \mathrm{context}) \) where \( \mathrm{context}=e_1,...,e_{n-1} \).

\paragraph{\e}
The \textbf{E}ntity-\textbf{G}uided \textbf{E}vent \textbf{L}anguage \textbf{M}odel generates the next event conditioned both on the context and one of its entities by learning the following probability distribution:
\[ \mathrm{P}(\mathrm{e_n} \mid \mathrm{context}, \mathrm{entity})\]
In this setting, the system learns to maximize the probability of the next event with respect to a specific entity from the context but without considering the specific role in which the entity appears in the generated event. The training instances in this case are (Context, Entity, Answer) tuples that are obtained by replacing the question with the entity mention present in the question in the training instances described in Section~\ref{sec:training}.

\paragraph{\q}
The \textbf{Q}uestion-\textbf{G}uided \textbf{E}vent \textbf{L}anguage \textbf{M}odel generates the next event conditioned both on the context and a question regarding one of the entities in the context and a \textbf{\textit{role}} the entity plays in the event. Here, the system learns the following probability distribution:
\[ \mathrm{P}(\mathrm{e_n} \mid \mathrm{context}, \mathrm{question})\]
The main difference with the entity-only system is the different roles an entity can have through the question's surface form.

\begin{table}
\centering
\small
\begin{tabular}{llc|cc}
\toprule

\multirow{4}{*}{\textbf{\makecell{Criteria}}} & \multirow{4}{*}{\textbf{System}} & \textbf{beam} & \multicolumn{2}{c}{\textbf{sampling}}  \\
\cline{3-5}
&&\textbf{\makecell{ Fail \%} \textcolor{red}{$\blacktriangledown$}} & \textbf{\makecell{ Fail \%} \textcolor{red}{$\blacktriangledown$}} & \textbf{\makecell{Avg \#  \\ samples }}\\
\midrule
\multirow{4}{*}{\makecell{Any\\ presence}} & ND &\phantom{0}{45.63} & - & - \\ & \base & \phantom{0}{79.29} & 26.05 & 17.21 \\  & \e & \phantom{0}{43.39}& \phantom{0}2.73 & \phantom{0}4.47 \\ & \q & \textbf{\phantom{0}{38.92}} & \textbf{\phantom{0}1.53} & \textbf{\phantom{0}3.51}\\
\midrule
\multirow{4}{*}{\makecell{Role \\ specific \\ presence}} & ND & \phantom{0}{69.84} & - & -\\ &  \base & \phantom{0}{85.55} & \phantom{0}{35.00} & 21.57\\  & \e & \phantom{0}{59.04} & \phantom{0}{6.34} & \phantom{0}{8.42}\\ & \q & \textbf{\phantom{0}43.03} & \textbf{\phantom{0}2.78}& \textbf{\phantom{0}4.54}\\
\bottomrule
\end{tabular}
\label{tab:control}
\caption{Controllability Assessment: Fail \% denotes the number of instances where the model fails to generate the specified entity. For sampling, we also show the average number of samples needed before an event with the specified entity is generated. 
ND denotes NeuroLogic decoding \cite{lu-etal-2021-neurologic}, a beam-search based controllable method with no sampling strategy.}
\end{table}

\subsection{Data Statistics}

We train all the models on the 2007 portion of the Annotated NYT corpus \cite{sandhaus2008new}. This subset contains a total of around 38k articles spanning over a 6 month period. Following the steps described earlier on these number of articles, we end up having over $700k$ instances (consisting of (context:$e_1,e_2,...,e_{t-1}$, question:$q$, answer:$e_t$) tuples) used to train the guided systems. For more details on the data statistics and experimental settings, please refer to the Appendix~\ref{append:stats} and ~\ref{append:exp}. The code will be released upon acceptance.


\section{Evaluation}\label{sec:eval}
\begin{table}
\centering
\renewcommand{\tabcolsep}{1.1mm}
\small
\begin{tabular}{lcc}
\toprule
{\textbf{System}} & {\textbf{Perplexity} \textcolor{red}{$\blacktriangledown$}} & \textbf{\makecell{NC Accuracy}\textcolor{blue}{$\blacktriangleup$}}\\

\midrule
\base & 24.64 & 46.2\%\\
\midrule
\e &  22.45 & 48.6\%\\ 
\e (margin) & 25.06 & 46.8\%\\ 
\midrule
\q & \textbf{22.11} & \textbf{49.3\%} \\ 
\q (margin) &  26.63 & 46.0\%\\
\bottomrule
\end{tabular}
\caption{Perplexity and the narrative cloze accuracy. Lower perplexity and higher accuracy is desirable.}
\label{tab:perplexity}
\end{table}

We assess the utility of the question-guided event language modeling in terms of four aspects: (i) Diversity: are they able to generate diverse sets of events that relate to a scenario? (ii) Control: do they generate events involving the specific entities in desired roles? (iii) Sequence modeling ability: how well can it predict observed events?, (iv) Interactive Utility: do users generate better sequences when using the model to collect events that fit a scenario?

\subsection{QGELM Improves Diversity}

We want event language models to generate diverse sequences covering different aspects of a scenario.
To assess diversity in generation, we first sample multiple sequences from the models. Given a context (starting with a context of length 1), we incrementally generate events by sampling one event at a time until we generate a sequence of a predefined length. We repeat this process to generate multiple sequences. We then measure the diversity of these sequences using Self-BLEU~\cite{zhu2018texygen}, which is the average of the BLEU scores~\cite{papineni2002bleu} when using one of the generated sequence as the output and the rest as references. 

%
First, we collected test instances that had a context length of one, which amounted to $938$ instances. The models were used to generate five sequences of lengths one through ten (i.e., five sequences of length one, five of length two and so on).
For \e and \q we randomly choose an entity/question to generate the next event. For each model, we then compute the Self-BLEU score of its five sequences of a specific length.
\hyperref[fig:diversity]{Figure 4} shows the average Self-BLEU over the test instances when generating sequences of different lengths. Lower Self-BLEU scores represent more diverse sequences.
Self-BLEU of question-guided outputs are lower compared to that of the other two models showing improved diversity. With longer sequences Self-BLEU increases for all models as there is more potential for overlap. However, the question-guided model retains higher levels of diversity compared to the rest. Standard event language modeling tends to cover the same types of events across different samples and to some extent conditioning on entities helps improve this to a small degree. Conditioning on the questions, however, yields significant gains in diversity showing promise for improved coverage of scenarios. 

\subsection{QGELM Controls for Entity Roles}
\begin{table}
\centering
\small
\begin{tabular}{lccc}
\hline
\textbf{Metric} &\textbf{\base} & \textbf{\q} & \textbf{change} \\
\hline
\# accepted events\textcolor{blue}{$\blacktriangleup$}  & 6.2 & 8.8 & $42\% \uparrow$\\
\# rejected steps\textcolor{red}{$\blacktriangledown$}  & 5.2 & 3.2 & $38\% \downarrow$ \\
\% rejected steps\textcolor{red}{$\blacktriangledown$} & 41.0 & 26.6 & $35\% \downarrow$\\
\# resamples\textcolor{red}{$\blacktriangledown$} & 4.9 & 3.2 & $35\% \downarrow$ \\
total steps\textcolor{blue}{$\blacktriangleup$} & 11.3 & 12.0 & $6\% \uparrow$\\
tree depth\textcolor{blue}{$\blacktriangleup$} & 5.8 & 8.8 & {$52\% \uparrow$}\\
\hline
\end{tabular}
\label{tab:interactive}
\caption{Quantitative analysis of schema generation using the ELM and QGELM models. With QGELM, users accepted more of its suggested events, rejected fewer steps, used fewer resamples for a given context, and produced longer event sequences. The higher the average the better a system is for metrics with \textcolor{blue}{$\blacktriangleup$} whereas lower values are desired for metrics with \textcolor{red}{$\blacktriangledown$}.}
\end{table}
\begin{figure*}
\centering
\includegraphics[width=\textwidth]{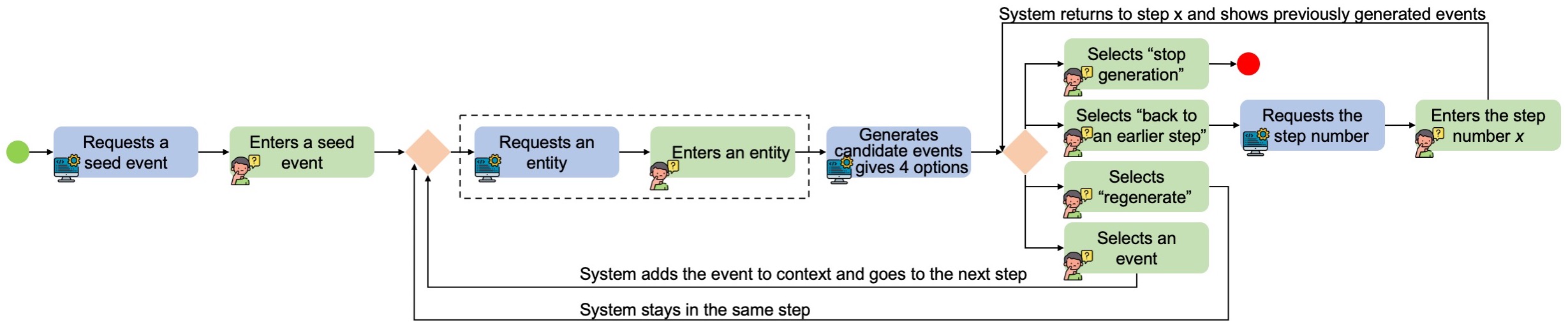}
\caption{The overview of the interactive schema generation tool. The dashed box is only used for \q.
The interaction starts by system asking for a seed event and the user entering an event. For \q, the user is asked for an entity of interest. This part is shown with a dashed box (this part is not needed for \base). The system then samples 4 events (If \q, it automatically creates questions for the given entity and generates two responses per question.) These 4 events will be presented to the user where they can select an option out of 4 choices. 1. Select one of the given events. 2. Ask the system to generate a new set of events. 3. Return to an earlier step to explore a different path by entering the step number and 4. Stop the generation for the given seed.}
\label{fig:tool}
\end{figure*}
To quantify controllability, we introduce a metric that measures how many times a system is capable of generating an event in which a specified entity of interest is present. We generate a fixed number of outputs from each model under two decoding strategies: sampling and beam decoding. We report the percentage of times the specified entity or a coreferent mention\footnote{A mention is considered coreferent to the input noun phrase if the mention and original noun phrase appear in the same coreference cluster extracted from the sampled sequence of events which includes the original context event.} of it fails to appear in the outputs. In addition to \base, we also use NeuroLogic decoding (ND)~\cite{lu-etal-2021-neurologic}, a state-of-the-art controllable generation system as a baseline.


%

The top block of \hyperref[tab:control]{Table 1} (Any presence) compares the models on the number of times they failed to generate an event with the specified entity within a beam of size $40$ or within $40$ sampling attempts (\%Fail), and the average number of events that had to be sampled to see the entity or its coreferent mention in the output when sampling (Avg \#samples). 
With beam decoding, \base performs the worst since it has no control over the generation. NeuroLogic decoding does a better job at searching for events that meet the entity constraints in the model's beam. \e which is trained to generate events with the given entity does better even with standard beam decoding. \q fares even better outperforming all methods by a significant margin. With sampling, ELM works better but still fails to produce events with the input entity in more than a quarter of the cases. Both entity and question guided models respond to the input control, almost always yielding events with the input entity and much earlier in the ranked list.

The bottom block of \hyperref[tab:control]{Table 1} (Role specific presence) compares the models when we are looking for events where the input entity is expected to appear in a specific role. We use the same dependency-based heuristics we described in Section~\ref{sec:training} for determining the role of an entity in a given event. With beam decoding,
NeuroLogic decoding (ND) is worse than \q with a larger margin since it can only be constrained to generate the entity but not in \textbf{\textit{specific}} position and role. 
The larger gap here shows the superiority of our question control codes to not only generate an entity but also to generate it in a specific role.
ELM fares even worse, with more failures and requiring even more samples to generate the entity in the specified role. Also, as expected, \e has a larger gap compared to \q which is trained to account for the roles in which the entities appear.




Note that the coreference resolution system and the dependency parser are not perfect and therefore our heuristics for deciding both when a entity is mentioned in an event and when it appears in a specific role can be faulty. A manual inspection of the outputs for 50 entities across all systems showed that the heuristic is 
more than $75\%$ accurate and the mistakes are uniform across the models.

\subsection{\q is a good event LM}
\begin{table*}
\centering
\small
\begin{tabular}{lcccccccc|c}
\hline
&\textbf{Crimes} & \textbf{Outbreak} & \textbf{Disaster} & \textbf{Kidnapping} & \textbf{Cyberattck} & \textbf{IED}  & \textbf{Shooting} & \textbf{Conflict} & \textbf{Overall}\\
\hline
\base & 10.71\ & 16.39 & 23.81 & 8.00 & 11.11 & 13.89 & 21.05 & \textbf{30.56} & 16.24\\
\q & \textbf{28.57} & \textbf{27.87} & \textbf{33.33} & \textbf{14.00} & \textbf{16.67} & \textbf{16.67}  & \textbf{23.68} & 25.00 & \textbf{23.22}\\
\hline
\end{tabular}
\label{tab:kairos}
\caption{Percentage of the overlap between system-generated events and manually curated schemas.
Overall, the \q system generates more diverse set of events, therefore having a higher recall (higher percentage of overlap) compared to the \base (for all domains except one). More details can be found in Appendix~\ref{append:kai}}
\end{table*}



How does question-guided training affect the raw ability to generate "standard" event sequences? Predicting observed events in a discourse can be seen as a downstream evaluation.
To assess this, we compare the perplexities as well as the narrative cloze task accuracy of the models using the event sequences we observe in the test set. The results are presented in \hyperref[tab:perplexity]{Table 2}. 
One way to turn the question guided-model into a standard event language model is to marginalize its probabilities for outputs over all possible questions we can ask at every step (marked as margin in the table): $P(e_n \mid C) = \sum_{q \in Q} P(q \mid C) P(e_n \mid q,C)$,
where we assume a uniform prior distribution $P(q \mid C)$ over questions that can be asked.
Similarly, for the entity guided model we can marginalize over the set of noun phrases in the most recent event in the context (this setting is similar to how we created the training instances). 
While this allows for fair comparison as a standard event language model, this is likely a lower bound for the model's ability for its intended use as a controllable model. 

\begin{table*}
\centering
\scriptsize
\begin{tabular}{p{1cm}p{4.5cm}p{9cm}}
\hline
\textbf{Domain} & \textbf{ELM} & \textbf{QGELM} \\
\hline
\multirow{8}{*}{\makecell{mass \\ shooting}}  & police evacuated the surrounding buildings after the shooting. \textcolor{teal}{the area resuscitated as soon as officials arrived. the area in search of survivors. a police helicopter carrying three officers fired at the shooting center. the police still searching for victims. the police found the bodies of four people. the bodies of two of them found in a second car.} & police evacuated the surrounding buildings after the shooting. \textcolor{teal}{he identified as a man.} \textcolor{violet}{a detective questioned the suspect.} \textcolor{teal}{the police identified the gunman. the officers heard noise from the building.} \textcolor{violet}{they spoke with him}. \textcolor{magenta}{he moved in an apartment that was recently renovated.} \textcolor{teal}{they worked on the scene to identify the gunman.} \textcolor{violet}{the officers involved in an investigation.} \textcolor{teal}{he notified about the shooting.} \textcolor{violet}{the police found no weapon in the apartment.} \textcolor{teal}{officers searched two buildings in the immediate area near the building.} \textcolor{blue}{the shooting began on saturday.} \textcolor{violet}{police officers questioned him about the shooting. The officer called victims at 6:45. the officer questioned again in the area.} \textcolor{blue}{the shooter shot two times. the shooting began with three or four shots of the gunman. he started firing slowly his gun}\\
\hline
\multirow{5}{*}{\makecell{kidnapping}} & the kidnapper ambushed the target. the kidnappers shot in the arm. \textcolor{red}{they} shot \textcolor{red}{the men}. \textcolor{red}{he} still not identified by the authorities. \textcolor{red}{the other men} also gunshot wounds. \textcolor{red}{the attack} left six people in critical condition and his condition. & the kidnapper ambushed the target. the kidnapping triggered by an act of rebellion. he left the scene. the target captured by a surprise attack. the kidnapping of the target essentially a symbolic step in the long struggle. \textcolor{red}{mr. seymour} to talk about the kidnapping. the police found a handgun. the hideout an informal area of community groups in rural part of town \\
\hline
\end{tabular}
\label{tab:examples}
\caption{Generated examples through human interaction. A mass shooting scenario in the first example, can include high level schemas such as \textcolor{magenta}{planning}, \textcolor{blue}{occurrence}, \textcolor{teal}{immediate response}, \textcolor{violet}{investigation}, etc. The \q covered more aspects compared to the \base. As for the \q, the entities can easily be tracked based on the questions asked, and therefore there are fewer ambiguous \textcolor{red}{red entities} (not clear what/who they are referring to.) compared to \base in the second example. More examples can be found in Appendix~\ref{append:example}.}
\end{table*}
We compute the average per token perplexities for the instances of the test set (including (C,Q,A) tuples) which are shown in \hyperref[tab:perplexity]{Table 2}.
Although the marginalized performance is lower because the model is forced to generate the event given suboptimal conditions, we see that with appropriate guidance (the question from the dataset itself), the perplexity of the true event is lower. Moreover, \q achieves the lowest perplexity in these settings, indicating that its control is most fine-grained and allows users the highest degree of control.
We also computed the accuracy of the cloze task in which each system has to correctly predict the gold output from a fixed set of events. Following ~\citet{weber2018hierarchical}, for each article from the test set, we randomly select an instance (context, question, answer) and then for each
instance we create 5 confounding events as answers by randomly
selecting instances from other documents. The task
would be then picking the correct option from the given six
choices (gold+confoundings). The results show that the accuracy trends align with the perplexity trends indicating that \q is comparable as a event language model to the baselines we build on.

\subsection{Interactive Schema Generation}\label{subsec:interactive}
Event language models can be used to generate event sequences that approximate a schematic description of how events typically happen in certain scenarios (i.e. event schemas)~\cite{weber2018event, weber2018hierarchical, pichotta2016learning}. 

\noindent \textbf{Task setup} \quad
We evaluate the utility of our new question-driven models when used in an interactive system, where a user collects a set of output events from the model that they think best describes a scenario. 
The overview of the system is depicted in \hyperref[fig:tool]{Figure 5}. We manually selected $35$ seeds from $8$ common domains and asked $7$ users (graduate students from NLP and non-NLP labs) to spend 4 minutes interacting with each system. They were asked to generate sequences of events for given seeds using \base and \q systems in a randomized order. 
For each scenario, the user is given a seed event and is tasked with collecting a set of events that \emph{best describe the scenario}.
At each step the user is presented with a set of generated events from the model and the user selects one of the events to add it to their collected set. The added events optionally become part of the conditioning context for more events. The user has the option to either regenerate events for the same context or go back and choose a subset of context from which to generate events. For the guided model, the events are generated by conditioning on questions. Since this is a timed practice, instead of asking the users to type in a full question, we only ask them to provide the system with an entity of interest. The system then automatically forms two questions (agentive, non-agentive) with the entity and outputs a mix of the events generated with all the questions. Additional details of this study including the settings, motivations for the timed version as well as instructions for the users are listed in Appendix~\ref{append:eval-all}.

At each step the user selects the best event that meets the following criteria: (i) \textit{Sensibleness}: whether the generated event is grammatically correct, sensible and easy to understand. (ii) \textit{Uniqueness}: events do not duplicate each other and describe different subevents. (iii) \textit{Relatedness}: events are related to the domain. (iv) \textit{Typicality}: the events are quite common for things in this domain and not too niche. For each event, the users make a binary judgment on whether each criteria is satisfied. If any of the criteria is not satisfied then the event is not selected. If no event meets all these criteria then the user either regenerates from the same context or moves to an earlier context.
Each user interaction with the system results in a sequence of events in the form of a tree, as users might have explored different paths by selecting different events at each step.
\hyperref[tab:examples]{Table 5} shows example outputs.

\noindent {\bf Analysis} \quad
\hyperref[tab:interactive]{Table 3} shows that with the \q based interactive system, users accept more of the system suggested events, which means that more events meet the criteria we set for good events. They ask for fewer resampling steps, require fewer returns to earlier steps and thus having fewer reject steps or wasted steps in their interaction. They also produce longer descriptions of the scenarios with higher tree-depth i.e. the length of the longest sequence they generated within a domain.


We further analyze the output generated using the interactive system to assess their utility in creating complex schematic knowledge. The seeds we use come from $8$ different domains relevant to the intelligence analysis community. For each domain we have access to manually curated schemas created by ten language experts in collaboration with the intelligence analysis experts over multiple days. An example of such schema is presented in \hyperref[tab:examples-manual]{Table 9} in Appendix~\ref{append:manual}.
We used these schemas as references and compared the percentage of overlap between system generated events selected by the users (within the four minute interaction) and the events in the reference schemas. The results in \hyperref[tab:kairos]{Table 4} 
show that both automatic systems can generate events that are expected to describe certain scenarios, however, the events generated by \q tend to have higher recall compared to the \base (in 7 out of 8 domains).
\section{Conclusion}\label{sec:conc}

Controlling event language models to generate events with respect to the participants is not trivial. We propose a simple yet effective question-guided approach that learns to generate events 
by not only conditioning on the events but also on specific entity-based questions of interest. Our empirical analysis shows that this approach can be used to generate more diverse sequences with better coverage and controllability allowing for better modeling of complex scenarios.

\section{Limitations}\label{sec:limit}

One of the limitations of our proposed approach is the coverage of the entity roles. We have used two broad categories of roles, mainly agentive (subject) and non-agentive (object) roles, however, there can be more fine-grained semantic roles for the participating entities in the events such as agent, patient, theme, manner, etc. Considering this taxonomy of semantic roles can lead to finer-grained questions which might lead to even richer descriptions of the scenarios.
Also, the human evaluation setting is limited since it is timed and users can not explore the models to their fullest extent. However, our analysis of the systems, when not timed, shows even a higher margin in terms of performance with the \q model. 
\section{Ethics Statement}\label{sec:ethics}

The models presented in the paper make use of the existing pretrained systems that train on large collections of data and are known to inherit biases that are existent in the training data. The event language models we train are also susceptible to these biases, which can result in generation of event sequences with these biases. 
\section*{Acknowledgments}
We thank the anonymous reviewers for their insightful
feedback and suggestions.
This material is based on research that is supported by the Air Force Research Laboratory (AFRL), DARPA, for the KAIROS program under agreement number FA8750-19-2-1003. The U.S. Government is authorized to reproduce and distribute reprints for Governmental purposes.

\bibliography{anthology,custom}
\bibliographystyle{acl_natbib}

\clearpage
\appendix
\section{Appendix}\label{appendix}

\subsection{Data Processing Details}\label{append:data}
\q uses the data in the form of (Context, Question, Answer) tuples. The details of the data processing is outlined in Algorithm ~\ref{alg:migrate}.
\begin{algorithm}[h]
\caption{Data Processing}
\label{alg:migrate}

\begin{algorithmic}[1]
\scriptsize
\Statex \textbf{Input:} $D$, Document, \textbf{Output:} $T$, List of (Context, Question, Answer) tuples

\State extract OpenIE event tuples $E$ from $D$
\State find co-referring clusters $Clusters$ from $D$ 
\State for $e_i$ in $E$:
\State    \quad find all nps $N_{e_i}$ 
\State    \quad for each $np$ in $N_{e_i}$: 
\State    \quad \quad generate $Q_{np}$
\State    \quad \quad for each $q_j$ in $Q_{np}$ for $np$ with role $r_j$
\State    \quad \quad \quad find an event $e_k$ $(k > i)$ with $np$ with role $r_j$
\State    \quad \quad \quad add $(e_1..e_i,q_j,e_k)$ to $T$

\end{algorithmic}
\end{algorithm}

We use AllenNLP \cite{gardner2018allennlp} for coreference resolution to find the clusters in a document and the spaCy \cite{Honnibal_spaCy_Industrial-strength_Natural_2020} dependency parser to identify all the noun phrases and their roles within an event. 

\subsubsection{Data Statistics}\label{append:stats}
We initially extract Open IE event tuples from  $37,924$ articles from the $2007$ portion of the NYT Annotated Corpus using Ollie \cite{schmitz2012open}. All the extracted events from a single document are concatenated to form a single event sequence for that document. Therefore, we end up having  $37,924$ event sequences with average length of $27$ events.

Then, we run the data processing algorithm on the extracted sequences to generate (Context, Question, Answer) tuples. \hyperref[tab:stats]{Table 6} shows the data statistics of the dataset.

\begin{table}[h]
    \centering
    \small
    \begin{tabular}{ccccc}
    \hline
        \textbf{split} & \textbf{Total} &\textbf{Q1} & \textbf{Q2} & \textbf{Q3} \\
        \hline
        All data & 762,004 & 466,757 & 188,300 & 106,947\\
        Train & 752,004 &460,573 &185,582 &105,579 \\
        Dev & 5,000 & 3,091 & 1,212 & 697\\
        Test & 5,000 & 3,093 & 1,236 & 671 \\
        \hline
        \makecell{Common \\ Test} & 18953 & 11,712 & 4,412 & 2,829\\
         \hline
    \end{tabular}
    \caption{Data Statistics of the (Context, Question, Answer) tuples for different types of questions. Q1 refers to \textit{what else happened?}, Q2 is \textit{what else did np do?} and Q3 is \textit{what else happened to np?}}
    \label{tab:stats}
\end{table}

\subsection{Experimental Settings}\label{append:exp}
\subsubsection{Input/Output format}
For the \base, the input will be the context and the output will be the next event. For \e and \q, the input will be the concatenation of the context and the entity/question, separated by [SEP] token. 
Since the input can be more than $512$ tokens (in case of long contexts), we need to truncate the input. Truncating the input from its end (for \e and \q) will result in input sequences without entity/question. To avoid this, we instead truncate from the beginning of the input sequence by removing the earlier events in the context. We remove events and not tokens from the context until its length is within the model input size.
The length of the output is also fixed at $50$ tokens.

\subsubsection{Systems Details}
Our systems finetune a pre-trained T5-base model and tokenizer with the implementation from Huggingface library \cite{wolf-etal-2020-transformers}. Adam optimizer \cite{kingma2014adam} is used with an initial learning rate of $6.25e-5$.

We use a batch size of $4$. Each training epoch takes almost $24$ hours to run. We use the dev set for early stopping. All the systems will converge after $3$ epochs.

\subsection{Evaluation}\label{append:eval-all}

\subsubsection{Interactive Evaluation Setup}\label{append:setup}
The details of the settings of the interactive evaluation are presented in \hyperref[tab:params]{Table 7}.

\subsubsection{Interactive Evaluation Design Choices}\label{append:eval}
Some design choices for this evaluation are constrained by practicalities that would ensure a fair comparison. We initially had used an untimed version where users could interact with the system indefinitely. However, we found that users spent different amounts of time working, making it near impossible to do fair comparisons across systems. Further, users found it difficult to retain focus over longer periods of time.

As for using an entity of interest instead of typing questions, typing questions at every stage induces a burden on the user and introduces variance because of typing speeds. But note that even though they only select an entity, the roles are used as part of the questions we generate for the entity. Half of the generated answers are with the entity in the subject role and the other half are in the object role. Users can pick whatever role they want to explore.

\begin{table}
\centering
\begin{tabular}{lc}
\hline
\textbf{Parameters} & \textbf{Values}\\
\hline
number of seed events &  35 \\
number of users & 7  \\
number of seeds per worker &  5 \\
allotted minutes & 4   \\
number of generated events per step &  4\\
number of domains & 8 \\
\hline
\multicolumn{2}{l}{\makecell[l]{\textbf{domains:} disease outbreak, cyberattack, ied, \\international conflict, kidnapping, disaster, \\ mass shooting, financial crimes}}
  \\
\hline
\end{tabular}
\label{tab:params}
\caption{The parameters of the human generation task.}
\end{table}

\begin{table}
\centering
\begin{tabular}{lcc}
\hline
\textbf{Domain} & \textbf{\# seeds} & \textbf{\# gold events}
\\
\hline
Crimes & 4 & 28\\ 
Outbreak & 6 & 61\\ 
Disaster & 5 & 42\\ 
Kidnapping & 4 & 50\\
Cyberattck & 3 & 54\\
IED  & 3 & 36\\
Shooting & 5 & 38\\
Conflict & 5 & 36\\
\hline
\end{tabular}
\label{tab:kairos-gold}
\caption{The statistics of the human evaluation in terms of number of seeds and the number of gold sequence events.}
\end{table}

\subsubsection{Instructions for Users}
Below, you can find the guidelines that were provided to the users prior to starting the task. 

\paragraph{User Manual} This study is aimed at evaluating the capabilities of the event language models in generating a sequence of events with their participating arguments that can be used to describe a scenario. For each scenario, a seed event will be given. Using the seed event, you can start generating the sequence incrementally by selecting the best event at each step based on the following set of criteria:

At each step select an event that is:
grammatical/understandable
non-redundant or unique: events do not duplicate each other and describe different subevents
on-topic: events are related to the domain and not unrelated
typical: the events are quite common for things in this domain and not too niche (for example, earthquake could cause a nuclear reactor to meltdown but it’s not common)

You will use two different systems to do this task. For one system you only need to select the events at each step while for the other, at each step you initially type an entity of interest and then pick the best event. 
This task will be timed, and you keep interacting with the system for 4 minutes, exploring different paths and entities. Once the time is over, the generation stops automatically. 

User Interface:
Once you run the commands, the system will load the pretrained models which will take a few seconds and then it will ask you to enter the seed event (The seeds will be given to you). Then you need to follow the prompts at each step for the allotted time. 
At each step there will be 4 actions you can take: 
Choose a preferred event generated at that step. You will be shown a set of events from which you can pick the best one according to the above criteria.
Regenerate events for the last step. If you feel none of the generated events satisfy the criteria, you can choose this option so that the system will generate a different set of events for this step. Please use this option if NONE of the generated events satisfy the criteria.
Choose an earlier step to return to. If you get stuck in one path and cannot generate events, you can choose to go back to an earlier step and continue the generation from a different path. Once you choose to return, the current set of events will be saved.
Stop generation for the given seed event. If you think the system has generated enough events to describe the scenario or if it is no longer generating good events, then you can decide to stop the generation even if the allotted time is not yet over.

Notes regarding the entities:
If you are asked for an entity, you can have the following things in mind:
You are given an initial set of entities that are relevant to the given scenario. You can pick entities from this set, think of other entities that might be relevant, pick entities from already generated events or just select ‘none’ if you cannot think of an entity.
You do not need to use all the given entities and you can choose the same entity if you think that is a main entity in the scenario or if you are interested in knowing more about that entity. You can generate a sequence which is centered around one specific entity. 
For instance, you can have an event sequence like this for the earthquake scenario where ‘earthquake’ is an argument in all the generated events: earthquake struck city, earthquake magnitude measured on scale, earthquake killed people, earthquake injured people, earthquake damaged buildings, earthquake disrupted services,...

\subsubsection{System-generated events VS Manually-curated schemas}\label{append:kai}
We tried to show the plausibility of the system-generated events by comparing the system outputs with schemas that are curated by a group of experts for different domains. 
We initially used a number of seed events from these domains and provided the users with these seeds to interact with the system. Then for all seeds from a single domain, we grouped all the generated events and measured the amount of overlap with human-written schemas.
The statistics of this experiment is presented in \hyperref[tab:kairos-gold]{Table 8}.
An event is considered to have an overlap with an event from the gold set if it either shares the exact predicate or a predicate with similar meaning. To do this, we provided a user with the list of system-generated events (not knowing which system this is coming from) and asked them to find the mappings between the gold set and the generated set. We then counted the number of events that are considered as overlapping with the gold events. 

\subsubsection{Manually curated schemas}\label{append:manual}
Real-life scenarios can be described with a sequence of events and their relations. Manually curated schemas represent the events that can unfold a scenario in a hierarchical structure.
The events in this structure can be either primitive or non-primitive, depending on whether they can be further expanded into additional events. We use the term ``schema'' to refer to  the non-primitive events. \hyperref[tab:examples-manual]{Table 9} shows an example of such schemas for the disaster domain. As can be seen, the events are represented in multiple levels. For the sake of the evaluation conducted in this work (to make the comparison of the system-generated sequences with curated schemas more compatible), we consider the flattened representations of the schemas which consists of concatenating all the primitive events into a single sequence of events. 

We also did the comparison on predicate level. The reason is that the system-generated events are instantiated events with specific arguments as the models are trained on news articles whereas the curated schemas are generalized forms of events. 

\begin{table*}
\centering
\small
\begin{tabular}{lcl}
\hline
\textbf{Schema name} &\textbf{Level} & \textbf{Events} \\
\hline
\multirow{9}{*}{natural disaster progression} & \multirow{9}{*}{0}& detection and tracking\textcolor{red}{*}\\
&&preparations\textcolor{red}{*}\\
&&damages\textcolor{red}{*}\\
&&immediate responses\textcolor{red}{*}\\
&&rescue organization\textcolor{red}{*} \\
&&rescue\textcolor{red}{*} \\
&&rescue outcome\textcolor{red}{*}\\
&&economic assistance\textcolor{red}{*}\\
&&rebuild damaged property\textcolor{red}{*}\\
\hline
\multirow{5}{*}{detection and tracking} & \multirow{5}{*}{1}&scientists detect warning signs\\
&&scientists track progress\\
&&scientists assess threat\\
&&scientists warn public\\
&&media broadcasts information\\
\hline
\multirow{3}{*}{preparations} & \multirow{3}{*}{1}&government announce order\\
&&people buy supplies\\
&&preparations outcome\textcolor{red}{*}\\
\hline
\multirow{6}{*}{damages} & \multirow{6}{*}{1}&disaster hurts person/people\\
&&disaster kills person/people\\
&&disaster destroys buildings\\
&&disaster damages infrastructure\\
&&disaster causes food/water shortage\\
&&disaster causes power outage\\
\hline
\multirow{6}{*}{immediate responses} & \multirow{6}{*}{1}&government counts deaths \\
&&government sends equipment\\
&&government estimates damage\\
&&government requests aid\\
&&person obtains aid\\
&&person sheltered\\
\hline
\multirow{3}{*}{rescue organization} & \multirow{3}{*}{1}&government hold session to plan rescue\\
&&government command rescue teams\\
&&rescue teams travel\\
\hline
\multirow{3}{*}{rescue} & \multirow{3}{*}{1}&government hold rescuers locate survivors\\
&&rescuers assess\\
&&rescuers triage survivors\\
\hline
\multirow{2}{*}{rescue outcome} & \multirow{2}{*}{1}&victims found dead\\
&&recovery\textcolor{red}{*}\\
\hline
\multirow{6}{*}{economic assistance} & \multirow{6}{*}{1}&government request help \\
&&government estimate loss\\
&&government make assistance policies\\
&&people apply for assistance\\
&&government gifts aid\\
&&government gives loans\\
\hline
\multirow{3}{*}{rebuild damaged property} & \multirow{3}{*}{1}&people contact insurers\\
&&people repair damage\\
&&people rebuild\\
\hline
\multirow{3}{*}{recovery} & \multirow{3}{*}{2}&rescuers save survivors\\
&&rescuers transport survivors\\
&&rescuers treat survivors\\
\hline
\multirow{2}{*}{preparations outcome} & \multirow{2}{*}{2}&evacuation\textcolor{red}{*}\\
&&people shelter in place\\
\hline
\multirow{3}{*}{evacuation} & \multirow{3}{*}{3}&people hear evacuation order\\
&&rescuer evacuate people\\
&&people leave affected area\\
\hline
\end{tabular}
\label{tab:examples-manual}
\caption{Manually curated schemas. Each domain is represented through a set of high-level schemas which in turn are expanded into multiple events. Here you can see the events of the disaster schema which are represented at different levels. Events with \textcolor{red}{*} are non-primitive events that are expanded.}
\end{table*}

\subsubsection{System generated event sequences}\label{append:example}
\hyperref[tab:examples2]{Table 10} shows examples of generated event sequences through users interaction with the interactive tool described in section ~\ref{subsec:interactive}.

\begin{table*}
\centering
\small
\begin{tabular}{p{2cm}p{2.5cm}p{1cm}p{8.5cm}}
\hline
\textbf{Domain} & \textbf{Seed} & \textbf{System} & \textbf{Generated event sequence}  \\
\hline
international conflict &country agreed to sign the accord. & \base & he country and several members of its foreign relations committee negotiated a peace agreement as part of the agreement. the agreement called for more cooperation. the terms of the agreement announced at the start of a yearlong talks. the talks took place in 2002. the deal still subject to approval. the agreement formally announced on saturday\\ && \q & peace take a major step. the accord a sign of change. the accord not necessarily end war. the agreement calls for a gradual suspension of all military combat and police operations. the accord also requests immediate withdrawal of troops\\
\hline
cyberattack & the cyberattack disrupted services & \base & the network crashed after a new attack. the virus affected internet connections. the virus infecting computers that have computer processors. two new attacks launched on thursday. the virus also infects networks that used the same basic operating system. two new attacks triggered an attack that affected 1 million internet users on tuesday. both attacks exposed another email containing information about a security company. one email message posted by an employee of a security company. security engineers investigating the e-mail messages. the other attack not affect the internet connection \\ && \q & the attack also infected some computer systems. at least 60 computers affected by the cyberattack. the attacks continued across an industrial network of 8. the attack started in friday. the attacks hit at least five businesses and four hospitals. the network failed in this attack. the attack attributed to an increase in unreported customer data from two years earlier to more widespread attacks. one problem the nature of these attacks. the attacks linked to an ''antiviral attack. the attackers focused to eliminate part of microsoft's core operating system that would detect the attack, and to prevent it from becoming more sophisticated. … … … the attacks aimed at businesses that are vulnerable to any type of attack. the attacks caused significant disruptions for internet service providers. they trying to improve the security. the attack involved a variety of devices. the hackers asked for information. the security department declined to comment about the attack.\\
\hline
disaster & the heavy flooding disrupted the services. & \base & the damage, which is estimated to cost about $\$2.5$ billion annually due to flooding and rainy weather largely a product of human error. the department of agriculture forecasts the flood. two million people killed in the flooding. a few roads halted on the west side. the power grids closed down in some areas. the flooding also damaged roads and bridges. the department of transportation and emergency management put forward a plan. most of the roads closed in new york. the department also announced at least 17 routes halted in the same area as they were closed on oct. 26\\ && \q &  the flooding resulted in a series of delays. the flooding also caused problems. the federal emergency management agency and the emergency maintenance team also monitoring damage from the flood. the flooding continued as far as tuesday. the flood already brought over from the south. the new york city fire department dispatched officers and engineers to help in the flooding. the flooding worsened on the west side. the flood occurred along the west side. about 3,000 people killed by the flooding. the flooding occurred north of manhattan around 1:55 p.m.. the water caused by another flooding a series of conditions. the flooding followed an investigation of a suspected bribery. the flooding still also affected the emergency management system. officials not expecting much rain until tuesday. the firefighters remained on manhattan. firefighters and crews sent out to assist residents with their tasks. the officials not identify a cause of the flooding. the flooding came after a storm that had swept up from manhattan. the rain caused by a river of soil. firefighters began an investigation into possible obstructions. the emergency crews working in all locations. they expected about eight feet of water to be there by this morning. the flood affected at least seven other areas. the flooding caused the agency to cut off access to the subways. the agency attributed much of the flooding to human error \\
\hline
\end{tabular}
\label{tab:examples2}
\caption{Generated examples through human interaction with the system.}
\end{table*}



\end{document}